\documentclass{article}
\pdfoutput=1

\date{}

\usepackage{enumitem}
\usepackage{spconf,amsmath,graphicx}
\usepackage{comment}
\usepackage{url}
\usepackage{hyperref}

\begin{document}

\onecolumn 

\begin{description}[leftmargin=2cm,style=multiline]

\item[\textbf{Citation}]{M. Prabhushankar, G. Kwon, D. Temel, and G. AlRegib, "Contrastive Explanations in Neural Networks," in \emph{IEEE International Conference on Image Processing (ICIP)}, Abu Dhabi, United Arab Emirates (UAE), Oct. 2020.}

\item[\textbf{Review}]{Date of publication: 25 Oct 2020}

\item[\textbf{Codes}]{\url{https://github.com/olivesgatech/Contrastive-Explanations}}

\item[\textbf{Copyright}]{\textcopyright 2020 IEEE. Personal use of this material is permitted. Permission from IEEE must be obtained for all other uses, in any current or future media, including reprinting/republishing this material for advertising or promotional purposes,
creating new collective works, for resale or redistribution to servers or lists, or reuse of any copyrighted component
of this work in other works. }

\item[\textbf{Contact}]{\href{mailto:mohit.p@gatech.edu}{mohit.p@gatech.edu}  OR \href{mailto:alregib@gatech.edu}{alregib@gatech.edu}\\ \url{http://ghassanalregib.com/} \\ }
\end{description}

\thispagestyle{empty}
\newpage
\clearpage
\setcounter{page}{1}

\twocolumn

\title{Contrastive Explanations in Neural Networks}
%
\name{Mohit Prabhushankar, Gukyeong Kwon, Dogancan Temel, and Ghassan AlRegib}
\address{OLIVES at the Center for Signal and Information Processing,\\ School of Electrical and Computer Engineering,\\ Georgia Institute of Technology, Atlanta, GA, 30332-0250\\ \{mohit.p, gukyeong.kwon, cantemel, alregib\}@gatech.edu}

\ninept
\maketitle
\begin{abstract}
Visual explanations are logical arguments based on visual features that justify the predictions made by neural networks. Current modes of visual explanations answer questions of the form \emph{`Why P?'}. These \emph{Why} questions operate under broad contexts thereby providing answers that are irrelevant in some cases. We propose to constrain these \emph{Why} questions based on some context $Q$ so that our explanations answer contrastive questions of the form \emph{`Why P, rather than Q?'}. In this paper, we formalize the structure of contrastive visual explanations for neural networks. We define contrast based on neural networks and propose a methodology to extract defined contrasts. We then use the extracted contrasts as a plug-in on top of existing \emph{`Why P?'} techniques, specifically Grad-CAM. We demonstrate their value in analyzing both networks and data in applications of large-scale recognition, fine-grained recognition, subsurface seismic analysis, and image quality assessment. 

\end{abstract}
\section{Introduction}
\label{sec:intro}

Explanations are a set of rationales used to understand the reasons behind a decision~\cite{kitcher1962scientific}. When these rationales are based on visual characteristics in a scene, the justifications used to understand the decision are termed as visual explanations~\cite{hendricks2016generating}. Visual explanations can be used as a means to interpret deep neural networks. While deep networks have surpassed human level performance in traditional computer vision tasks like recognition~\cite{he2016deep}, their lack of transparency in decision making has presented obstacles to their widespread adoption. We first formalize the structure of visual explanations to motivate the need for the proposed contrastive explanations. Hempel and Oppenheim~\cite{hempel1948studies} were the first to provide formal structure to explanations~\cite{wilkenfeld2014functional}. They argued that explanations are like proofs in a logical system~\cite{keil2006explanation} and that explanations elucidate decisions of hitherto un-interpretable systems. Typically, explanations involve an answer to structured questions of the form \textit{`Why P?'}, where $P$ refers to any decision. For instance, in recognition algorithms, $P$ refers to the predicted class. In image quality assessment, $P$ refers to the estimated quality. \emph{Why}-questions are generally thought of to be \textit{causal-like} in their explanations~\cite{koura1988approach}. In this paper, we refer to them as visual causal explanations for simplicity. Note that these visual causal explanations do not allow causal inference as described by~\cite{pearl2009causal}.

\begin{figure}
\begin{center}
\minipage{0.49\textwidth}%
  \includegraphics[width=\linewidth]{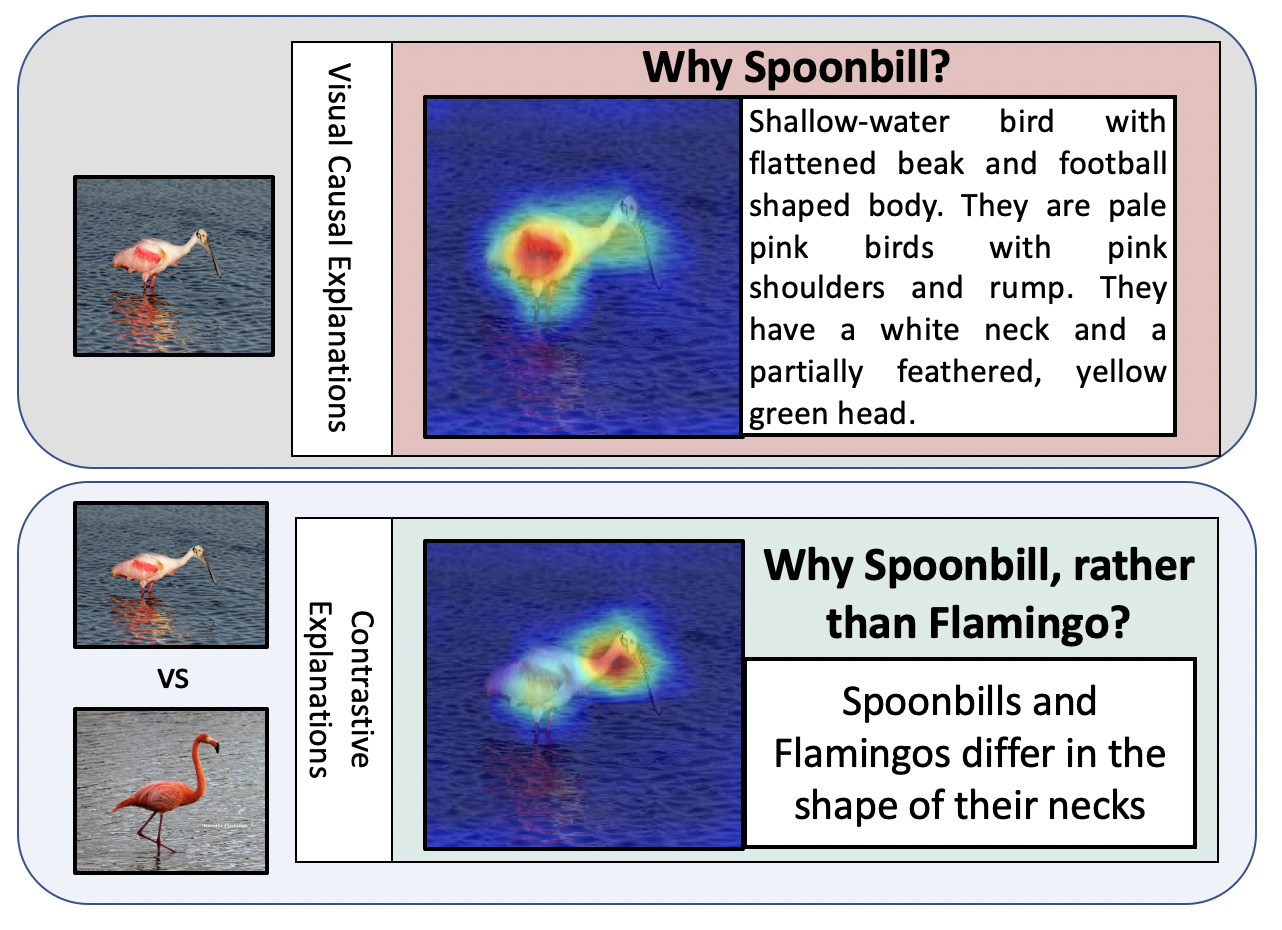}
\endminipage
\vspace{-4mm}
\caption{The visual explanation to \textit{Why Spoonbill?} is answered through Grad-CAM. The proposed contrastive explanatory method explains \textit{Why Spoonbill, rather than Flamingo?} by highlighting the neck region in the same input image. Figure best viewed in color.}\label{fig:concept}
\vspace{-5mm}
\end{center}
\end{figure}

Consider an example shown in Fig.~\ref{fig:concept} where we classify between two birds - a spoonbill, and a flamingo. Given a spoonbill, a trained neural network classifies the input correctly as a spoonbill. A visual explanation of its decision generally assumes the form of a heat map that is overlaid on the image. In the visual explanations shown in Fig.~\ref{fig:concept}, the red regions answer the posed question. If the posed question takes the form of \textit{`Why Spoonbill?'}, then the regions corresponding to the body shape and color of the spoonbill are highlighted. Such an explanation is based on features that describe a Spoonbill irrespective of the context. Instead of \textit{`Why Spoonbill?'}, if the posed question were \textit{`Why Spoonbill, rather than Flamingo?'}, then the visual explanation points to the most contrastive features between the two birds, which in this case is the neck of the Spoonbill. Flamingos have a longer S-shaped neck not prevalent in Spoonbills. The answers to such \emph{`Why P, rather than Q?'} questions are \emph{contrastive explanations} where $Q$ is the contrast.

The question of \emph{`Why P, rather than Q?} provides context to the answer and hence relevance~\cite{mayes2001theories}. In some cases, such context can be more descriptive for interpretability. For instance, in autonomous driving applications that recognize traffic signs, knowing why a particular traffic sign was chosen over another is informative in contexts of analyzing decisions in case of accidents. Similarly, in the application of image quality assessment where an algorithm predicts the score of an image as $0.25$, knowing \textit{`Why 0.25, rather than 0.5?'} or \textit{`Why 0.25, rather than 1?'} can be beneficial to analyze both the image and the method itself. In applications like seismic analysis where geophysicists interpret subsurface images, visualizing \textit{`Why fault, rather than salt dome?'} can help evaluating the model, thereby increasing the trust in such systems. In this paper, we set the framework for contrastive explanations in neural networks. More specifically, we modify existing \textit{`Why P?'} explanatory systems like Grad-CAM to obtain contrastive explanations in Section~\ref{sec:contrast}. We show its usage in varied applications in Section~\ref{sec:experiments}. We then conclude in Section~\ref{sec:conclusion}.

\vspace{-3mm}
\section{Background and Related Works}
\label{sec:background}
\vspace{-1mm}
We propose to constrain \textit{'Why P?'} explanatory techniques by providing them context and relevance to obtain \textit{'Why P, rather than Q?'} techniques. In this section, we describe the existing \textit{'Why P?'} techniques and lay the mathematical foundations of neural networks.

\noindent\textbf{Background:} Consider an $L$ layered classification network $f()$, trained to differentiate between $N$ classes. Given an input image $x$, $f()$ provides a probability score $y$ of dimensions $N\times 1$ where each element in $y$ corresponds to the probability of $x$ belonging to one of $N$ classes. The predicted class $P$ of image $x$ is the index of the maximum element in $y$ i.e. $P = \operatorname{arg\,max}_i  f(x) \forall i \in [1,N]$. During training, an empirical loss $J(P,y',\theta)$ is minimized where $y'$ is the ground truth and $\theta$ is the network weights and bias parameters. Backpropagation~\cite{rumelhart1986learning} minimizes the loss $J()$ by traversing along the parameter space using gradients $\frac{\partial J}{\partial \theta}$. These gradients represent the change in network required to predict $y'$ instead of $P$. Note that for a regression network $f()$, the above mathematical foundations remain consistent with $P$ and $y'$ being continuous rather than discrete. 

\noindent\textbf{\emph{Why P?} Explanations:} A number of proposed techniques attempt to visually explain \textit{`Why P?'}. The authors in~\cite{simonyan2013deep} backpropagate the class probabilities $y$ and show that the obtained gradients are descriptive of the class $i, i \in [1,N]$. They also show that notions of all $N$ classes are learned in the network. Grad-CAM~\cite{selvaraju2017grad} localizes the \textit{`Why P?'} parts of the image by backpropagating a class-weighted one-hot vector and multiplying the averaged resultant gradients as importance-weights on activation maps produced by the input image $x$. In this paper, we combine the gradient's role as a loss minimizer in backpropagation, the existence of notion of classes within neural nets from~\cite{simonyan2013deep}, and~\cite{selvaraju2017grad}'s importance-weighing of activation maps to obtain our contrastive explanations. The authors in~\cite{goyal2019counterfactual} tackle contrast through counterfactuals. They change the input image, based on a distractor image, to change its prediction from $P$ to $Q$. In this paper, we use the existence of notion of classes to provide contrastive explanations without the need for changes to the input images.

\begin{figure}[!t]
\begin{center}
\minipage{0.4\textwidth}%
  \includegraphics[width=\linewidth]{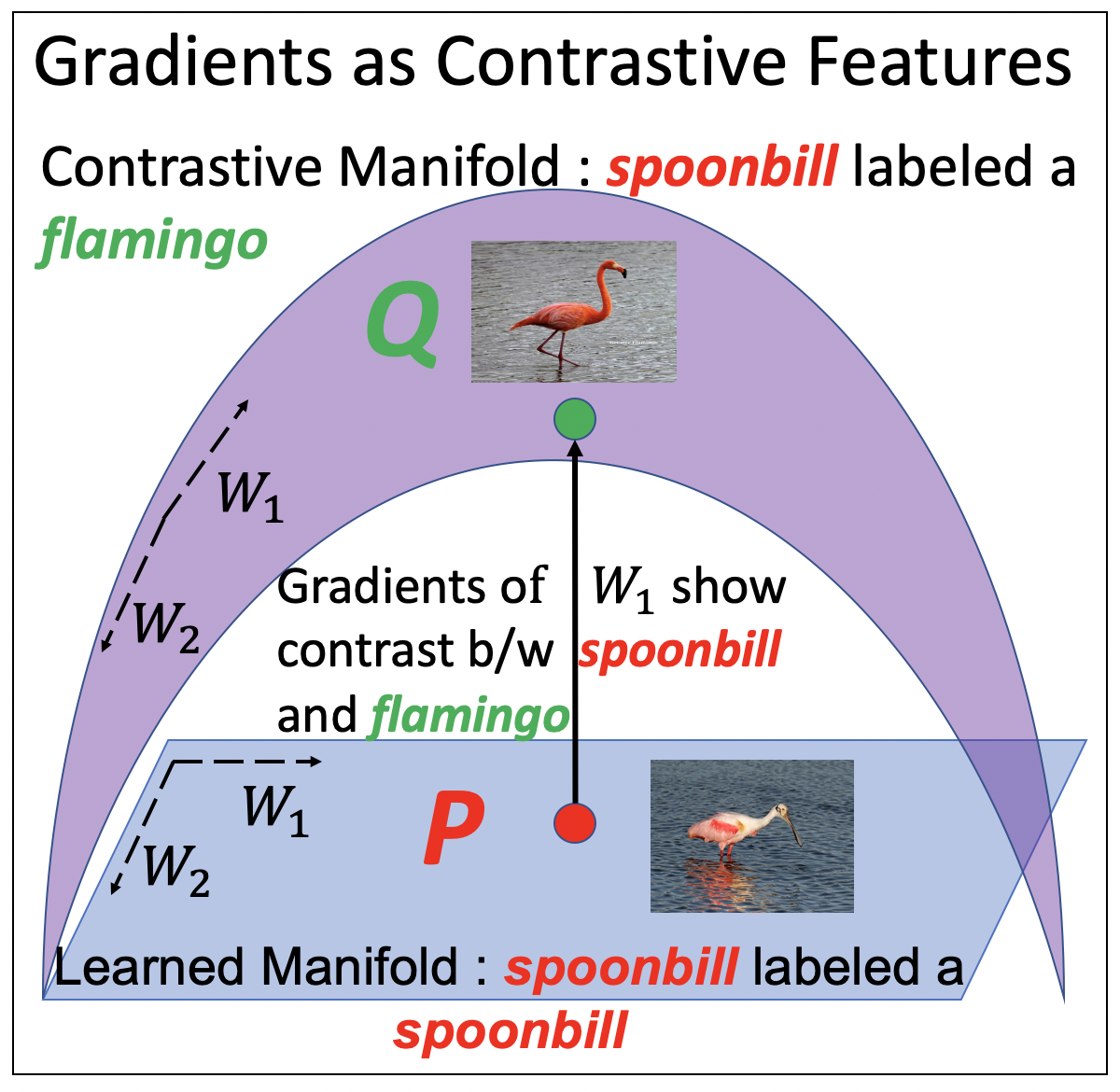}
\endminipage
\vspace{-3mm}
\caption{$P$ manifold in blue is the learned manifold that recognizes a spoonbill as a spoonbill. $Q$ is the contrastive manifold where spoonbill is classified as a fl. Change between the two is termed contrast.}\label{fig:contrast}
\vspace{-3mm}
\end{center}
\end{figure}
\vspace{-3mm}
\section{Contrastive Explanation Generation}
\label{sec:contrast}
\vspace{-1mm}
We define contrast and provide a methodology to generate then from neural networks. We embed contrast in existing \textit{`Why P?'} explanations, specifically Grad-CAM, to obtain contrastive explanations.

\vspace{-3mm}
\subsection{Contrast in Neural Networks}
\label{subsec:contrast}
\vspace{-1mm}
In visual space, we define contrast as the perceived difference between two known quantities. In this paper, we assume that the knowledge of the two quantities is provided by a neural network. For instance, in Fig.~\ref{fig:concept}, a neural network is trained to recognize both spoonbills and flamingos as separate classes. Thus, the network has access to the discriminative knowledge that separates the two classes. This knowledge is stored in the network's weight and bias parameters, termed as $W$ and $b$ respectively. These parameters span a manifold where the given image $x$ belongs to a class $i, i \in [1,N]$. A toy classification example is shown in Fig.~\ref{fig:contrast} where a learned manifold is visualized in blue. On the learned manifold, a spoonbill is classified as a spoonbill. A hypothetical contrastive manifold is shown in purple that differs from the blue manifold in that it recognizes a spoonbill as a flamingo. The same figure holds for regression networks, where the manifolds exist in a continuous space rather than discrete space. In terms of neural network representation space, contrast is the difference between the manifolds that predict $x$ as $P$ and $x$ as $Q$. In this paper, instead of directly measuring the difference between learned and contrastive manifolds, we measure the change required to obtain the contrastive manifold from the learned manifold. We use gradients to measure this change. Usage of gradients to characterize model change in not new. The authors in~\cite{kwon2019distorted} used gradients with respect to weights to characterize distortions for sparse and variational autoencoders. Fisher Vectors use gradients to characterize the change that data creates within networks~\cite{jaakkola1999exploiting} which were extended to classify images~\cite{sanchez2013image}.

\begin{figure*}[!t]
\begin{center}
\minipage{1\textwidth}%
  \includegraphics[width=\linewidth]{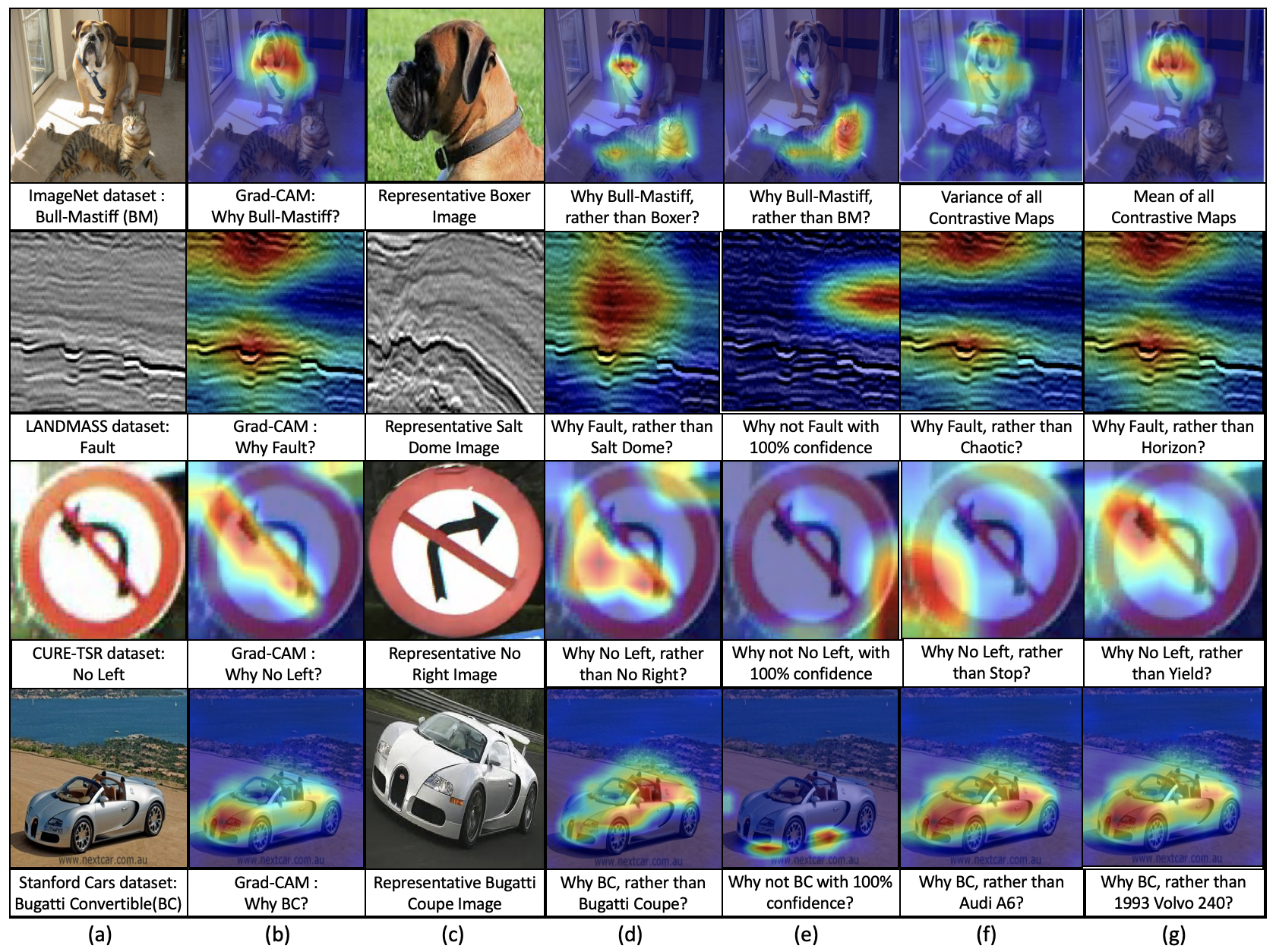}
\endminipage
\vspace{-3mm}
\caption{Contrastive Explanations (CE) on Recognition. (a) Input $x$. (b) Grad-CAM of $x$ for predicted class $P$. (c) Representative image of nearest class $Q$. (d) CE for class $Q$. (e) CE when $P = Q$. (f) and (g) CE for random classes. Figure best viewed in color.}\label{fig:Recognition}
\vspace{-3mm}
\end{center}
\end{figure*}

We extract contrast for class $Q$ when an image $x$ is predicted as $P$ by backpropagating a loss between $P$ and $Q$. Hence, for a loss function $J()$, contrast is proportional to $J(P,Q,\theta)$, where $\theta$ are the network parameters. For a contrastive class $Q$, contrast is $\frac{\partial J(P,Q,\theta)}{\partial \theta}$.  Note that $J()$ is a measure of contrastivity between $P$ and $Q$. In this paper, we choose $J()$ to be cross-entropy for recognition networks and mean square error for regression networks. The contrastive class $Q$ can belong to any one of the learned classes, i.e. $Q \in [1,N]$. Moreover, if $f()$ is a regression network such as in image quality assessment, $Q$ can take on any value in the range of $f()$.

\vspace{-3mm}
\subsection{Contrastive Explanations}
\label{subsec:contrast_explanations}
\vspace{-1mm}
For a network $f()$ that predicts $P$ for a given image, the gradients obtained from Sec.~\ref{subsec:contrast} represent \textit{`Why P, rather than Q?'}. They provide the difference between the predicted class or variable $P$ and the network's notion of class or variable $Q$. These features can now be plugged into any \textit{`Why P?'} based methods to obtain visual explanations. In this paper, we use Grad-CAM~\cite{selvaraju2017grad} to showcase our contrastive explanations. Essentially, the obtained contrastive gradients are backpropagated to the last convolutional layer to obtain $K$ gradient maps, where $K$ is the number of channels in that layer. These gradients are average pooled and the obtained $K\times 1$ vector is used importance-weights across the activation maps in the last convolution layer. These weighted activation maps are mean pooled and resized back to the original image dimensions to obtain contrastive masks. The contrastive masks are overlaid as heat maps and shown.

\vspace{-3mm}
\section{Applications}
\label{sec:experiments}
\vspace{-3mm}
In this section, we consider two applications : recognition and image quality assessment (IQA). Visualizing contrast between classes is instructive in interpreting whether a network has truly learned the differences between classes in recognition. In IQA, visualizing contrast can help us to both localize the exact regions of quality degradation as well as quantify degradation based on scores. For recognition $P$ and $Q$ are discrete classes while for image quality assessment, $P$ and $Q$ are continuous and can take values between $[0,1]$. 

\begin{figure*}[!htb]
\begin{center}
\minipage{1\textwidth}%
  \includegraphics[width=\linewidth]{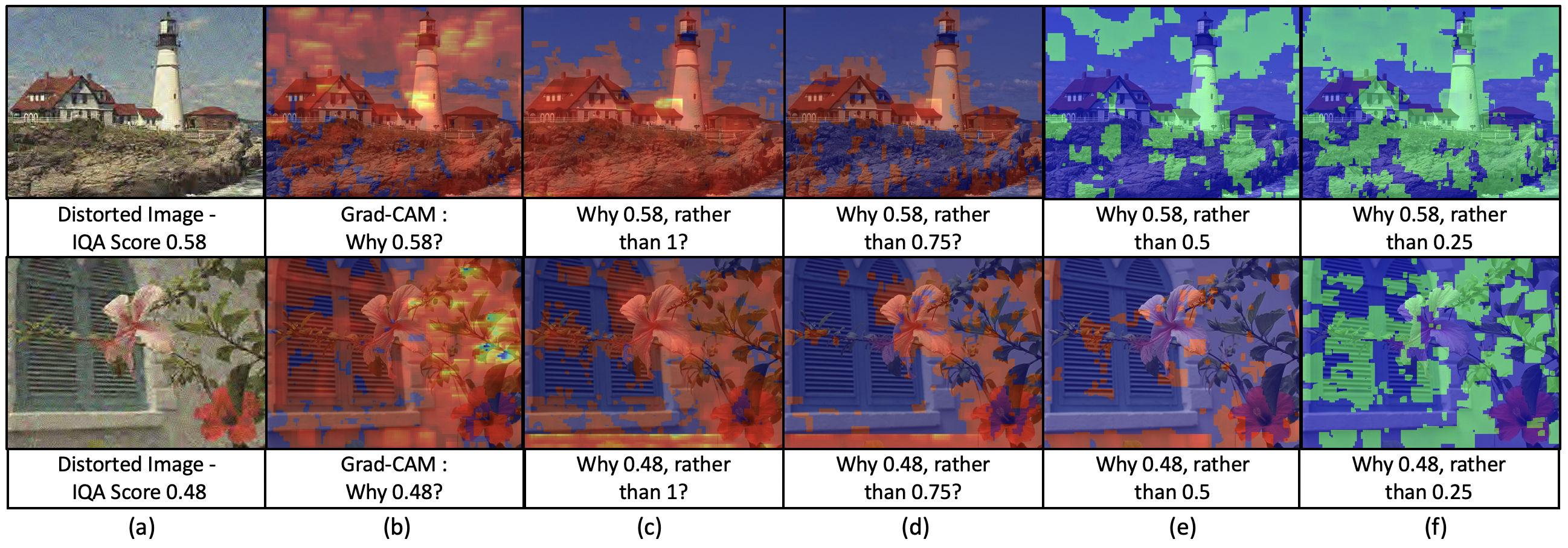}
\endminipage
\vspace{-3mm}
\caption{(a) Distorted images (b) Grad-CAM (c)-(f) Contrastive explanations to questions shown below each image. Best viewed in color.}\label{fig:IQA}
\vspace{-3mm}
\end{center}
\end{figure*}

\vspace{-3mm}
\subsection{Recognition}
\label{subsec:recognition}
\vspace{-1mm}
\noindent\textbf{Experiment} In this section, we consider contrastive explanations on large scale and fine-grained recognition. Large-scale dataset, like ImageNet~\cite{ILSVRC15}, consists of a wide variety of classes. Fine-grained recognition is the subordinate categorization of similar objects such as different types of birds, and cars among themselves~\cite{yang2012unsupervised}. We consider Stanford Cars~\cite{KrauseStarkDengFei-Fei_3DRR2013}, subsurface imaging based LANDMASS~\cite{alaudah2015curvelet}, and traffic sign recognition CURE-TSR~\cite{temel2017cure, Temel2018_SPM, temel2019traffic}, datasets for fine-grained recognition. We use PyTorch's ImageNet pretrained models including AlexNet~\cite{krizhevsky2012imagenet}, SqueezeNet~\cite{iandola2016squeezenet}, VGG-$16,19$~\cite{simonyan2014very}, ResNet-$18,34,50,101$~\cite{he2016deep}, and DenseNet-$161$~\cite{huang2017densely} to obtain contrastive explanations on ImageNet, Stanford Cars, and LANDMASS datasets. Specifically, on Stanford Cars and LANDMASS, we replace and train the final fully connected layer with the requisite number of classes - $196$ for Cars and $4$ for LANDMASS datasets. For CURE-TSR, we use the trained network provided by the authors in~\cite{temel2017cure} to extract contrastive explanations. The results of randomly selected images from the fine-grained datasets and the cat-dog image used in Grad-CAM paper~\cite{selvaraju2017grad} are shown in Fig.~\ref{fig:Recognition}. Similar to~\cite{selvaraju2017grad}, we show results on VGG-16. Note that the contrastive explanations are a function of the network $f()$. Hence, based on how good the explanations are, we can rank different networks. However, in this paper, we focus only on demonstrating the need and descriptive capability of contrastive explanations.

\noindent\textbf{Analysis:} ImageNet has $1000$ classes. Hence, for every image there are $999$ contrasts with a wide range of class options. This creates potential contrastive questions like \textit{'Why bull-mastiff, rather than golf ball?'}. The potential visual explanation to such a question lies in the face of the dog. Similarly, when asked \textit{'Why bull-mastiff, rather than minibus?'}, the potential visual explanation is the face of the dog. Hence, the contrastive explanations between a majority of $999$ classes to an image $x$ belonging to a class $P$ is the same. This is illustrated in row $1$ of Fig.~\ref{fig:Recognition}. The Grad-CAM image of an input predicted bull-mastiff is shown in Fig.~\ref{fig:Recognition}b. The face of the dog is highlighted. We calculate all $999$ contrastive maps of the input image and show their variance and mean images in Fig.~\ref{fig:Recognition}f. and Fig.~\ref{fig:Recognition}g. respectively. The variance map is boosted by a factor of $5$ for readability. For classes that are visually similar to bull-mastiff, like that of a boxer, the contrastive explanations indicate the semantic differences between the two classes. This is illustrated in Fig.~\ref{fig:Recognition}d. where the contrastive explanations show that there is a difference in the snout between the expected image and the network's notion of a boxer illustrated in Fig.~\ref{fig:Recognition}c. When $P$ is the same as $Q$, the contrastivity reduces to $J(P,P,\theta)$. This is the same as the loss function during backpropagation and hence the contrastive gradients act as training gradients in that their purpose is to confidently predict $P$. Hence, the contrastive explanation in this case highlights those regions in the image that is limiting the network from predicting $P$ with $100\%$ confidence. This is shown in Fig.~\ref{fig:Recognition}e. where the cat confuses the network $f()$ and is highlighted in red.

We show results on three fine-grained recognition datasets in rows $2,3,4$ of Fig.~\ref{fig:Recognition}. Note that the contrastive explanations in this case are descriptive between different classes. Representative images from similar classes are shown in Fig.~\ref{fig:Recognition}c. and their corresponding contrastive explanations are visualized in Fig.~\ref{fig:Recognition}d. The contrastive explanations track the fault when asked to contrast with a salt dome (Row 2 Fig.~\ref{fig:Recognition}d.), highlight the missing bottom part of the arrow in the no right turn image (Row 3 Fig.~\ref{fig:Recognition}d.), and highlight the roof when differentiating between the convertible and the coupe (Row 4 Fig.~\ref{fig:Recognition}d.). Other explanations to random classes are also shown. The input Bugatti Veyron's sloping hood is sufficiently different from that of the boxy hood of both the Audi and the Volvo that it is highlighted. 

\vspace{-3mm}
\subsection{Image Quality Assessment}
\label{subsec:IQA}
\vspace{-1mm}

\noindent\textbf{Experiment:} Image Quality Assessment (IQA) is the objective estimation of the subjective quality of an image~\cite{temel2016unique}. In this section, we analyze a trained end-to-end full-reference metric DIQaM-FR~\cite{8063957}. Given a pristine image and its distorted version, the pretrained network from~\cite{8063957} provides a quality score, $P$, to the distorted image. We then use MSE loss function as $J()$ and a real number $Q \in [0,1]$ to calculate the contrastive gradients. Contrastive explanations of $Q$ values including $0.25, 0.5, 0.75,\text{ and } 1$ along with Grad-CAM results are shown in Fig.~\ref{fig:IQA}. Both the lighthouse and flower images are distorted using lossy compression and are taken from TID 2013 dataset~\cite{ponomarenko2015image}. Note that the network analyzes the results patchwise. To not filter out the results of individual patches, we visualize un-normalized results in Fig.~\ref{fig:IQA}. In this implementation, while $f()$ takes non-overlapping patches to estimate quality, we use overlapping patches with stride $4$ to obtain smoother visualization. Note that the green and red colored regions within the images are the explanations to the contrastive questions shown below each image.

\noindent\textbf{Analysis:} Fig.~\ref{fig:IQA}b. shows that Grad-CAM essentially highlights the entire image. This indicates that the network estimates the quality $P$ based on the whole image. However, the contrastive explanations are indicative of where in the image, the network assigns quality scores. Fig.~\ref{fig:IQA}c. and d. show the regions within the distorted images that prevent the network from estimating $P = 1\text{ or } 0.75$. According to the obtained visualizations, the estimated quality is primarily due to the distortions within the foreground portions of the image as opposed to the background. This falls inline with previous works in IQA that argue that distortions in the more salient foreground or edge features cause a larger drop in perceptual quality than that in color or background~\cite{prabhushankar2017ms}\cite{chandler2013seven}. Also, the results when $Q = 0.75$ are different from their $Q = 1$ counterparts. For instance, the network estimates the quality of the distorted lighthouse image to be $0.58$. The results in \emph{`Why 0.58, rather than 0.75?'} show that the distortion in the lighthouse decreases the quality from $0.75$ to $0.58$. Similarly, the results from \emph{`Why 0.58, rather than 1?'} show that because of distortions in the lighthouse as well as the cliff and parts of the background sky, the estimation is $0.58$ rather than $1$. These results help us in further understanding the notion of perceptual quality within $f()$. When $Q < P$, the contrastive explanations describe why a higher rating is chosen. It can be seen that the network considers both the foreground and background to estimate a higher quality than $0.25$. We intentionally choose different visualization color maps when $P > Q$ vs when $P < Q$ to effectively analyze these scenarios.

\vspace{-3mm}
\section{Conclusion}
\label{sec:conclusion}
\vspace{-1mm}
In this paper, we formalized the structure of contrastive explanations. We also provided a methodology to extract contrasts from networks and use them as plug-in techniques on top of existing visual explanatory methods. We demonstrated the use of contrastive explanations in fine-grained recognition to differentiate between subordinate classes. We also demonstrated the ability of contrastive explanations to analyze distorted data and provide answers to contrastive questions in image quality assessment. 

\bibliographystyle{IEEEbib.bst}
\bibliography{references}

\begin{thebibliography}{10}

\bibitem{kitcher1962scientific}
Philip Kitcher and Wesley~C Salmon,
\newblock {\em Scientific explanation}, vol.~13,
\newblock U of Minnesota Press, 1962.

\bibitem{hendricks2016generating}
Lisa~Anne Hendricks, Zeynep Akata, Marcus Rohrbach, Jeff Donahue, Bernt
  Schiele, and Trevor Darrell,
\newblock ``Generating visual explanations,''
\newblock in {\em European Conference on Computer Vision}. Springer, 2016, pp.
  3--19.

\bibitem{he2016deep}
Kaiming He, Xiangyu Zhang, Shaoqing Ren, and Jian Sun,
\newblock ``Deep residual learning for image recognition,''
\newblock in {\em Proceedings of the IEEE conference on computer vision and
  pattern recognition}, 2016, pp. 770--778.

\bibitem{hempel1948studies}
Carl~G Hempel and Paul Oppenheim,
\newblock ``Studies in the logic of explanation,''
\newblock {\em Philosophy of science}, vol. 15, no. 2, pp. 135--175, 1948.

\bibitem{wilkenfeld2014functional}
Daniel~A Wilkenfeld,
\newblock ``Functional explaining: A new approach to the philosophy of
  explanation,''
\newblock {\em Synthese}, vol. 191, no. 14, pp. 3367--3391, 2014.

\bibitem{keil2006explanation}
Frank~C Keil,
\newblock ``Explanation and understanding,''
\newblock {\em Annu. Rev. Psychol.}, vol. 57, pp. 227--254, 2006.

\bibitem{koura1988approach}
Antti Koura,
\newblock ``An approach to why-questions,''
\newblock {\em Synthese}, vol. 74, no. 2, pp. 191--206, 1988.

\bibitem{pearl2009causal}
Judea Pearl et~al.,
\newblock ``Causal inference in statistics: An overview,''
\newblock {\em Statistics surveys}, vol. 3, pp. 96--146, 2009.

\bibitem{mayes2001theories}
G~Randolph Mayes,
\newblock ``Theories of explanation,''
\newblock 2001.

\bibitem{rumelhart1986learning}
David~E Rumelhart, Geoffrey~E Hinton, and Ronald~J Williams,
\newblock ``Learning representations by back-propagating errors,''
\newblock {\em nature}, vol. 323, no. 6088, pp. 533--536, 1986.

\bibitem{simonyan2013deep}
Karen Simonyan, Andrea Vedaldi, and Andrew Zisserman,
\newblock ``Deep inside convolutional networks: Visualising image
  classification models and saliency maps,''
\newblock {\em arXiv preprint arXiv:1312.6034}, 2013.

\bibitem{selvaraju2017grad}
Ramprasaath~R Selvaraju, Michael Cogswell, Abhishek Das, Ramakrishna Vedantam,
  Devi Parikh, and Dhruv Batra,
\newblock ``Grad-cam: Visual explanations from deep networks via gradient-based
  localization,''
\newblock in {\em Proceedings of the IEEE international conference on computer
  vision}, 2017, pp. 618--626.

\bibitem{goyal2019counterfactual}
Yash Goyal, Ziyan Wu, Jan Ernst, Dhruv Batra, Devi Parikh, and Stefan Lee,
\newblock ``Counterfactual visual explanations,''
\newblock {\em arXiv preprint arXiv:1904.07451}, 2019.

\bibitem{kwon2019distorted}
Gukyeong Kwon, Mohit Prabhushankar, Dogancan Temel, and Ghassan AlRegib,
\newblock ``Distorted representation space characterization through
  backpropagated gradients,''
\newblock in {\em 2019 IEEE International Conference on Image Processing
  (ICIP)}. IEEE, 2019, pp. 2651--2655.

\bibitem{jaakkola1999exploiting}
Tommi Jaakkola and David Haussler,
\newblock ``Exploiting generative models in discriminative classifiers,''
\newblock in {\em Advances in neural information processing systems}, 1999, pp.
  487--493.

\bibitem{sanchez2013image}
Jorge S{\'a}nchez, Florent Perronnin, Thomas Mensink, and Jakob Verbeek,
\newblock ``Image classification with the fisher vector: Theory and practice,''
\newblock {\em International journal of computer vision}, vol. 105, no. 3, pp.
  222--245, 2013.

\bibitem{ILSVRC15}
Olga Russakovsky, Jia Deng, Hao Su, Jonathan Krause, Sanjeev Satheesh, Sean Ma,
  Zhiheng Huang, Andrej Karpathy, Aditya Khosla, Michael Bernstein,
  Alexander~C. Berg, and Li~Fei-Fei,
\newblock ``{ImageNet Large Scale Visual Recognition Challenge},''
\newblock {\em International Journal of Computer Vision (IJCV)}, vol. 115, no.
  3, pp. 211--252, 2015.

\bibitem{yang2012unsupervised}
Shulin Yang, Liefeng Bo, Jue Wang, and Linda~G Shapiro,
\newblock ``Unsupervised template learning for fine-grained object
  recognition,''
\newblock in {\em Advances in neural information processing systems}, 2012, pp.
  3122--3130.

\bibitem{KrauseStarkDengFei-Fei_3DRR2013}
Jonathan Krause, Michael Stark, Jia Deng, and Li~Fei-Fei,
\newblock ``3d object representations for fine-grained categorization,''
\newblock in {\em 4th International IEEE Workshop on 3D Representation and
  Recognition (3dRR-13)}, Sydney, Australia, 2013.

\bibitem{alaudah2015curvelet}
Yazeed Alaudah and Ghassan AlRegib,
\newblock ``A curvelet-based distance measure for seismic images,''
\newblock in {\em 2015 IEEE International Conference on Image Processing
  (ICIP)}. IEEE, 2015, pp. 4200--4204.

\bibitem{temel2017cure}
Dogancan Temel, Gukyeong Kwon, Mohit Prabhushankar, and Ghassan AlRegib,
\newblock ``Cure-tsr: Challenging unreal and real environments for traffic sign
  recognition,''
\newblock {\em arXiv preprint arXiv:1712.02463}, 2017.

\bibitem{Temel2018_SPM}
D.~Temel and G.~AlRegib,
\newblock ``Traffic signs in the wild: Highlights from the ieee video and image
  processing cup 2017 student competition [sp competitions],''
\newblock {\em IEEE Sig. Proc. Mag.}, vol. 35, no. 2, pp. 154--161, 2018.

\bibitem{temel2019traffic}
D.~Temel, M.~Chen, and G.~AlRegib,
\newblock ``Traffic sign detection under challenging conditions: A deeper look
  into performance variations and spectral characteristics,''
\newblock {\em IEEE Transactions on Intelligent Transportation Systems}, pp.
  1--11, 2019.

\bibitem{krizhevsky2012imagenet}
Alex Krizhevsky, Ilya Sutskever, and Geoffrey~E Hinton,
\newblock ``Imagenet classification with deep convolutional neural networks,''
\newblock in {\em Advances in neural information processing systems}, 2012, pp.
  1097--1105.

\bibitem{iandola2016squeezenet}
Forrest~N Iandola, Song Han, Matthew~W Moskewicz, Khalid Ashraf, William~J
  Dally, and Kurt Keutzer,
\newblock ``Squeezenet: Alexnet-level accuracy with 50x fewer parameters and<
  0.5 mb model size,''
\newblock {\em arXiv preprint arXiv:1602.07360}, 2016.

\bibitem{simonyan2014very}
Karen Simonyan and Andrew Zisserman,
\newblock ``Very deep convolutional networks for large-scale image
  recognition,''
\newblock {\em arXiv preprint arXiv:1409.1556}, 2014.

\bibitem{huang2017densely}
Gao Huang, Zhuang Liu, Laurens Van Der~Maaten, and Kilian~Q Weinberger,
\newblock ``Densely connected convolutional networks,''
\newblock in {\em Proceedings of the IEEE conference on computer vision and
  pattern recognition}, 2017, pp. 4700--4708.

\bibitem{temel2016unique}
Dogancan Temel, Mohit Prabhushankar, and Ghassan AlRegib,
\newblock ``Unique: Unsupervised image quality estimation,''
\newblock {\em IEEE signal processing letters}, vol. 23, no. 10, pp.
  1414--1418, 2016.

\bibitem{8063957}
S.~{Bosse}, D.~{Maniry}, K.~{Müller}, T.~{Wiegand}, and W.~{Samek},
\newblock ``Deep neural networks for no-reference and full-reference image
  quality assessment,''
\newblock {\em IEEE Transactions on Image Processing}, vol. 27, no. 1, pp.
  206--219, Jan 2018.

\bibitem{ponomarenko2015image}
Nikolay Ponomarenko, Lina Jin, Oleg Ieremeiev, Vladimir Lukin, Karen
  Egiazarian, Jaakko Astola, Benoit Vozel, Kacem Chehdi, Marco Carli, Federica
  Battisti, et~al.,
\newblock ``Image database tid2013: Peculiarities, results and perspectives,''
\newblock {\em Signal Processing: Image Communication}, vol. 30, pp. 57--77,
  2015.

\bibitem{prabhushankar2017ms}
Mohit Prabhushankar, Dogancan Temel, and Ghassan AlRegib,
\newblock ``Ms-unique: Multi-model and sharpness-weighted unsupervised image
  quality estimation,''
\newblock {\em Electronic Imaging}, vol. 2017, no. 12, pp. 30--35, 2017.

\bibitem{chandler2013seven}
Damon~M Chandler,
\newblock ``Seven challenges in image quality assessment: past, present, and
  future research,''
\newblock {\em ISRN Signal Processing}, vol. 2013, 2013.

\end{thebibliography}

\end{document}